\documentclass[conference]{IEEEtran}
\IEEEoverridecommandlockouts
\usepackage{cite}
\usepackage{comment}
\usepackage{amsmath,amssymb,amsfonts}
\usepackage{algorithmic}
\usepackage{graphicx}
\usepackage{pdfpages}
\usepackage{subfigure}
\usepackage{textcomp}
\usepackage{hyperref}
\usepackage{xcolor}
\def\BibTeX{{\rm B\kern-.05em{\sc i\kern-.025em b}\kern-.08em
    T\kern-.1667em\lower.7ex\hbox{E}\kern-.125emX}}
\begin{document}

\title{Shuffle Vision Transformer: Lightweight, Fast and Efficient Recognition of Driver’s Facial Expression
\thanks{Thanks to the EUNICE Alliance                                            for the financial support.}
}
 \author{
    \IEEEauthorblockN{Ibtissam Saadi\textsuperscript{1,2}, Douglas W. Cunningham\textsuperscript{2}, Taleb-ahmed Abdelmalik\textsuperscript{1}, Abdenour Hadid\textsuperscript{3}, Yassin El Hillali\textsuperscript{1}}
    \IEEEauthorblockA{\textsuperscript{1}\textit{Laboratory of IEMN, CNRS, Centrale Lille, UMR 8520}, \textit{Univ. Polytechnique Hauts-de-France, F-59313}, France \\
    \textsuperscript{2}\textit{Faculty of Graphical Systems, Univ. BTU Cottbus-Senftenberg, Cottbus, Germany} \\
     \textsuperscript{3}\textit{Sorbonne Center for Artificial Intelligence, Sorbonne University Abu Dhabi, Abu Dhabi, UAE}\\
    \IEEEauthorblockA{Email: \{ibtissam.saadi, abdelmalik.taleb-ahmed, yassin.elhillali\}@uphf.fr, douglas.cunningham@b-tu.de, abdenour.hadid@ieee.org}\\
}
 }
\maketitle

\begin{abstract}
Existing methods for driver's facial expression recognition (DFER) are often computationally intensive, rendering them unsuitable for real-time applications. In this work, we introduce a novel transfer learning-based dual architecture, named "ShuffViT-DFER," which elegantly combines computational efficiency and accuracy. This is achieved by harnessing the strengths of two lightweight and efficient models using convolutional neural network (CNN) and vision transformers (ViT). We efficiently fuse the extracted features to enhance the performance of the model in accurately recognizing the facial expressions of the driver. Our experimental results on two benchmarking and public datasets, KMU-FED and KDEF, highlight the validity of our proposed method for real-time application with superior performance when compared to state-of-the-art methods. 

\end{abstract}


\section{Introduction}
Human factors are responsible for a significant percentage of traffic road accidents~\cite{website1}. For this reason, there has been an increasing interest on driver's facial expression recognition as a potential solution to improve road safety. Autonomous vehicles and Advanced Driver Assistance Systems (ADAS) both incorporate this feature, which enable recognizing and comprehending the emotional state of the driver. As a result, the systems are able to make well-informed decisions, which help to create a road environment that is safer and more effective.

In this context, several research works have focused on the development of techniques for the recognition of driver's facial expressions as for example ~\cite{patil2019driver},~\cite{tauqeer2022driver},~\cite{sahoo2022deep1}, and~\cite{yang2023robust}. However, most of these attempts were faced with the challenge of operating in real-time while accurately recognizing the driver's emotional state in a real-world driving environment. This challenge usually involves a range of factors, such as non-frontal driver-head position, occlusions, and variation in lighting condition. While some approaches, especially those based on deep learning (e.g.~\cite{mustafa2022manta},~\cite{wu2018accurate}) perform better compared to traditional machine learning-based techniques (e.g.~\cite{patil2019driver},~\cite{azman2019real}), they require large amounts of data and significant computational resources for model training, making them less attractive in real-time applications. Nevertheless, some recent works have attempted to overcome the computational challenge by utilizing lightweight models designed to efficiently operate and with low latency at the cost of lower performance, as observed in~\cite{jeong2020lightweight},~\cite{shang2023driver}.

Our work introduces a novel transfer learning-based approach that combines performance and computational efficiency by leveraging lightweight and efficient models, making it suitable for embedded systems and real-time applications. The main contributions of our present work can be summarized as follows:

\begin{itemize}
\item We introduce ShuffViT-DFER, a novel lightweight, and efficient approach for driver's facial expression recognition. 
\item We leverage the strengths of ShuffleNet V2~\cite{ma2018shufflenet} and EfficientViT~\cite{liu2023efficientvit} architectures, exploiting features from both models with a new classification scheme, achieving accurate and fast recognition. 
\item We demonstrate that dual-architecture transfer learning allows to efficiently capture subtle facial cues and expressions with limited data while maintaining real-time processing.
\item We explore the use of Efficient ViT approach in drivers' facial expression recognition by utilizing Grid Search to find the optimal hyper-parameters.
\item We conduct extensive experiments and evaluations on two benchmarking and publicly available databases, obtaining interesting performance compared to the state-of-the-art.
\item To support the principle of reproducible research, we share the code with the research community for comparison and future extensions at: \href{https://github.com/Ibtissam-SAADI/ShuffViT-DFER}{https://github.com/Ibtissam-SAADI/ShuffViT-DFER}.

\end{itemize}

\begin{figure*}[!h]
    \centering
    \includegraphics[width=0.722\linewidth]{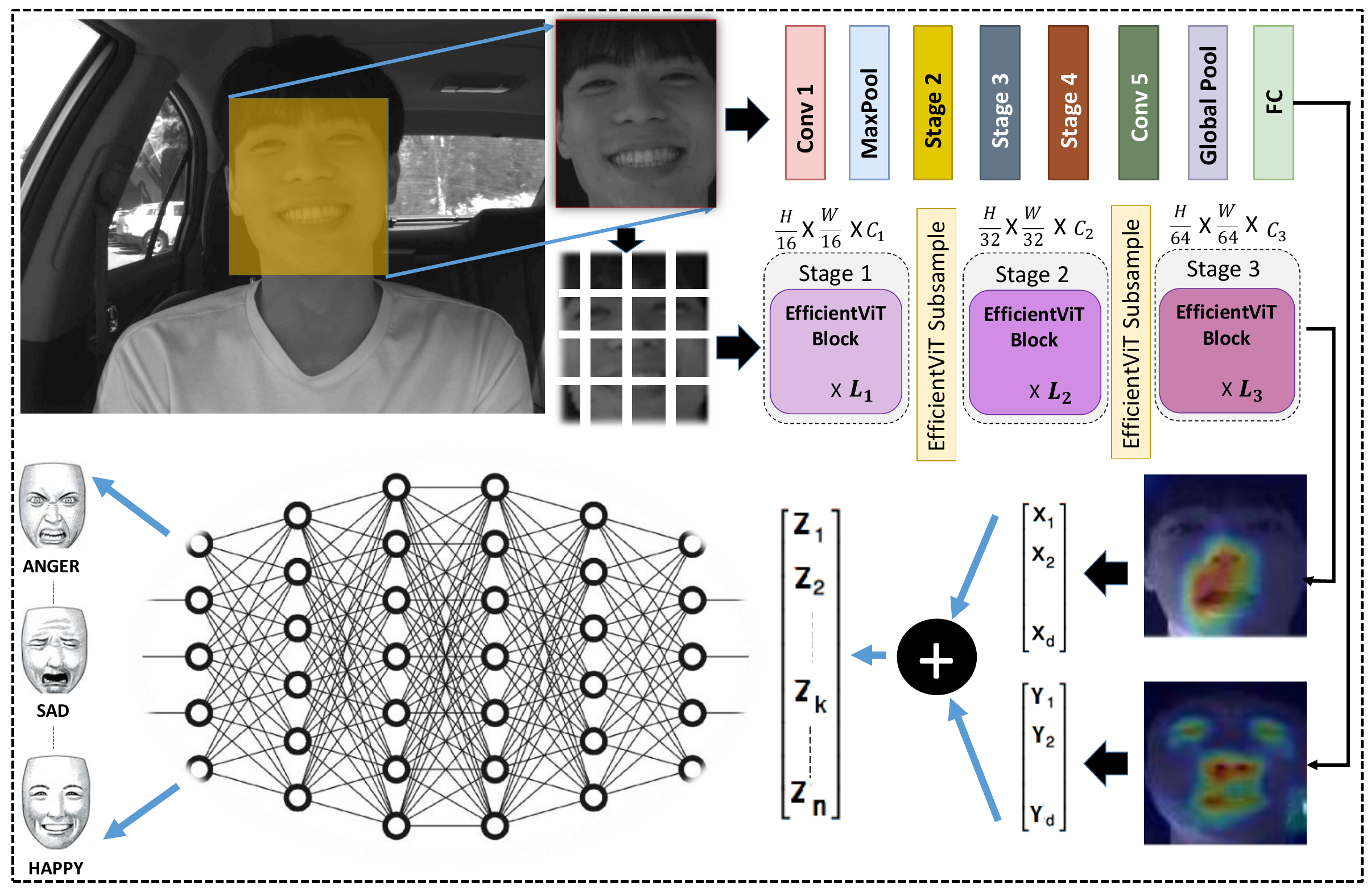}
    \caption{The workflow of our proposed model for driver facial expression recognition.}
    \label{fig:pip1}
\end{figure*} 

\section{Proposed Model: ShuffViT-DFER}\label{sec:PM}
The pipeline of our proposed architecture is illustrated in Figure~\ref{fig:pip1}. The input to our model consists of a cropped face, detected using Multi-Task Cascaded Convolutional Networks (MTCNN)~\cite{zhang2016joint}. Subsequently, the detected faces undergo data augmentation for enhancing the size of the training set. Features are extracted using two pretrained models namely ShuffleNet V2~\cite{ma2018shufflenet} and EfficientViT-M2~\cite{liu2023efficientvit}. The extracted features are fed to a classifier for accurate expression recognition.

\subsection{Preprocessing}
The faces are first detected using MTCNN~\cite{zhang2016joint}. Then, they are cropped and resized into 224$\times$224 pixels. Then, data augmentation is used by applying Random Horizontal Flip, Random Rotation, ColorJitter, Random Affine, and Gaussian Blur. The goal is to alleviate the overfitting issues related with small FER datasets. Finally, face normalization is performed.

\subsection{Lightweight and efficient feature extraction}
We extract features from both ShuffleNet V2 and EfficientViT-M2 models, employing transfer learning to address the limited data constraints and harness the strengths of both models for effective feature extraction, thereby enhancing classification accuracy. ShuffleNet V2, a lightweight CNN architecture, is known for its high computational efficiency and remarkable accuracy. However, it has limitations in capturing complex hierarchical features compared to deeper networks, which are needed for distinguishing subtle facial expressions. The outputs of ShuffleNet V2 model can be written as: $X = (X_1, X_2, \ldots, X_d)$. To enrich the extracted features, we also consider the High-Speed ViT family, specifically EfficientViT-M2. EfficientViT-M2 has indeed shown to be effective in encoding both local and global information, including more complex features, while utilizing limited computational resources. Its outputs can be written as: $Y = (Y_1, Y_2, \ldots, Y_d)$. We fuse the features extracted from both models into a single feature vector to enhance recognition accuracy and maintain trade-off between speed and accuracy. As a result we obtain: $Z = (Z_1, Z_2, \ldots, Z_k, \ldots, Z_n)$ which can be expressed as: $ Z = X \oplus Y $.
\subsection{Classification}
For accurate classification, we consider three fully connected layers, facilitating linear transformations to capture additional features. The inclusion of two batch normalization layers serves to stabilize and accelerate the training process. Additionally, two ReLU activation functions are applied to introduce non-linearity and capture complex relationships within the data. To mitigate overfitting, we integrate two dropout layers, randomly disabling a portion of input units during training.

 \begin{figure}[!t]
    \centering
    \includegraphics[width=0.36\textwidth]{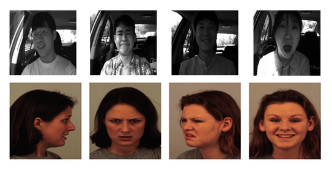}
    \caption{Samples from the data used in the experiments. Top: KMU-FED dataset; Bottom: KDEF dataset.}
    \label{fig:dts}
\end{figure}
\section{Experimental Analysis}\label{sec:Ex}
Our model was implemented using the open-source PyTorch framework, on an NVIDIA GPU device, specifically the Quadro RTX 5000 with 16 GB of RAM. We conducted the experiments using two publicly available datasets namely KMU-FED and KDEF (See Figure~\ref{fig:dts} for some face samples).

\begingroup
\begin{figure*}[t]
\centering
\subfigure[KMU-FED(10-fold)]{\includegraphics[width=0.483\linewidth]{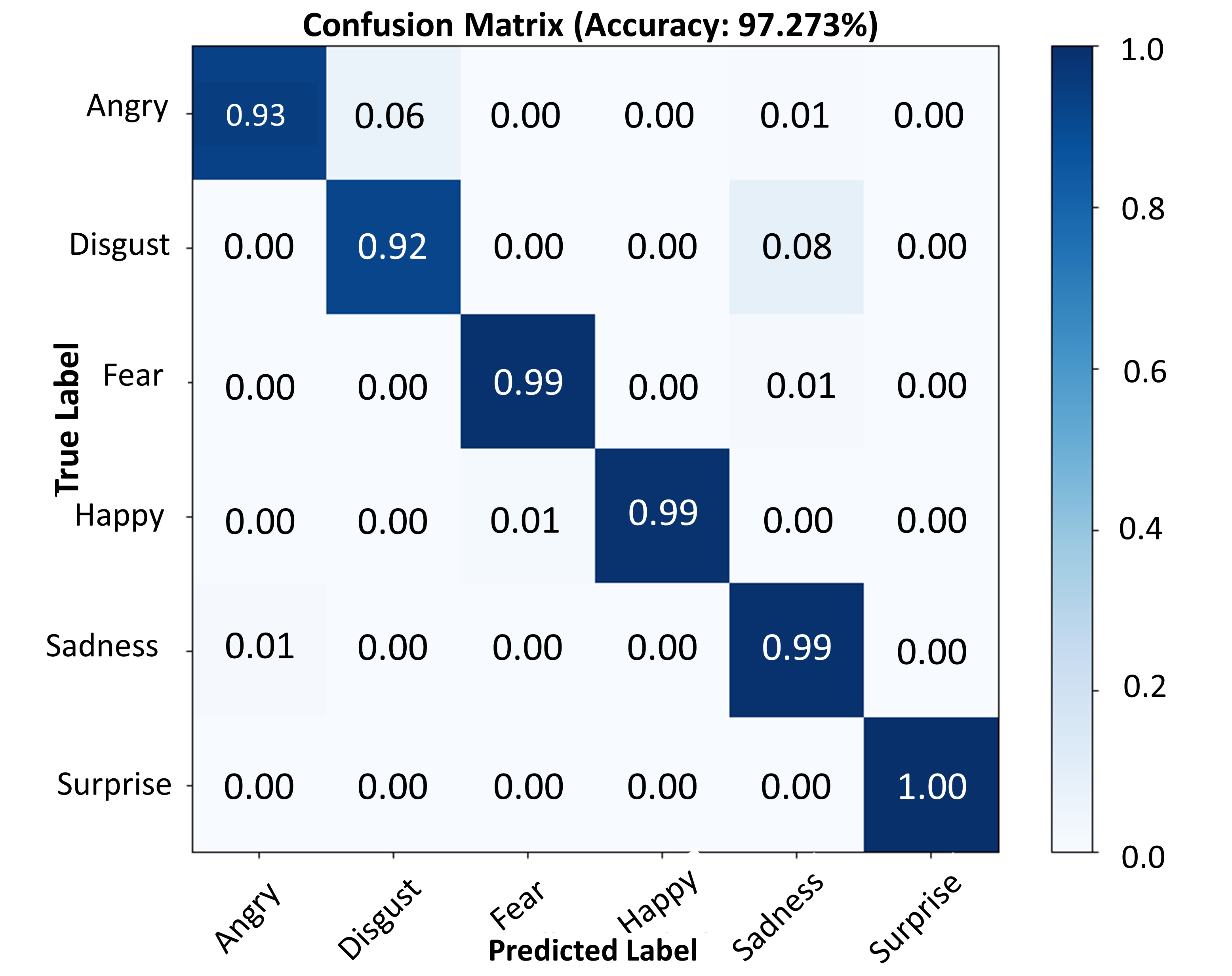}}\label{fig:1a}
\subfigure[KDEF] {\includegraphics[width=0.483\linewidth]{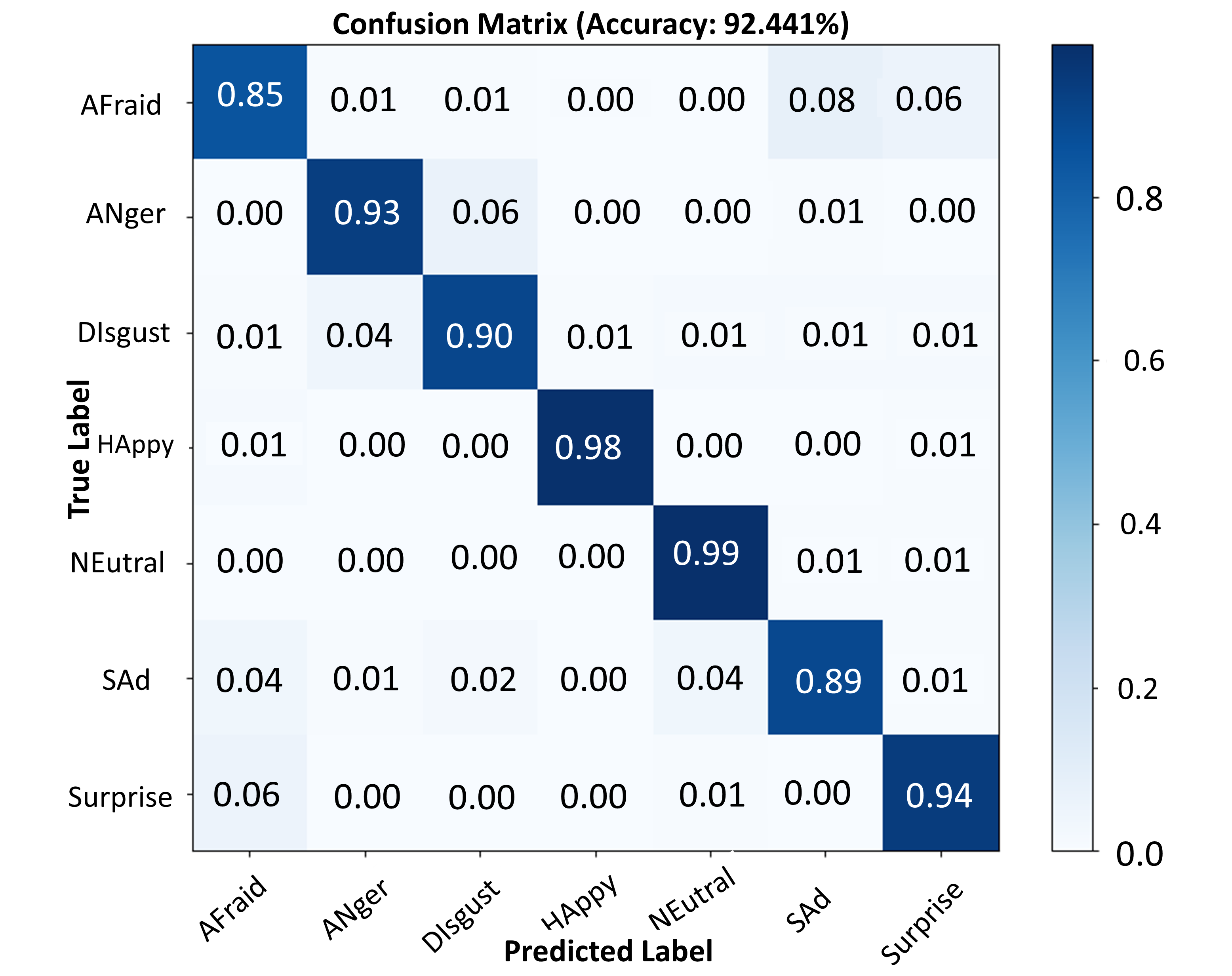}}\label{fig:1d}
\caption{The results of our model shown as confusion matrices of KMU-FED and KDEF datasets.}
\label{fig:cnx}
\end{figure*}
\endgroup
\subsection{Experimental Data}
\paragraph{\bf KMU-FED} To evaluate the efficiency of our approach in real-world driving scenarios, we first used KMU-FED (Keimyung University Facial Expression of Drivers)~\cite{jeong2018driver} dataset, which is captured in real driving environment. The database has a total of 1106 images from 12 subjects, with labels for the six basic emotions. The dataset includes different lighting variations and partial occlusions caused by hair or sunglasses. For comprehensive evaluation, we considered both 10-fold and 5-fold cross-validation protocol and a train--test ratio of 80\%--20\%.

\paragraph{\bf KDEF} In addition to KMU-FED dataset, we also considered the Karolinska Directed Emotional Face (KDEF) dataset~\cite{lundqvist1998karolinska}, comprising 4900 images of human emotional facial expressions captured from 35 male and 35 female subjects at five different angles: $-90^\circ$, $-45^\circ$, $0^\circ$, $45^\circ$, and $90^\circ$, and without any accessories, makeup, or glasses. It includes seven different emotions (afraid, angry, disgust, happy, neutral, sad, and surprise). For a fair comparison with previous works, we divided the KDEF dataset using a train--test ratio of 80\%--20\%.

\subsection{Experimental Setup}
We utilized the Grid Search technique to determine the optimal hyper-parameters on both datasets. For the KMU-FED dataset, we used a batch size of 128 and the model was optimized using the Adaptive Moment Estimation (Adam) optimizer with a fixed learning rate of 0.001. The training process extended over 90 epochs, employing the cross-entropy loss function. For the KDEF dataset, a batch size of 32 was employed, and a learning rate of 0.0001 was utilized. Similar to the KMU-FED dataset, the Adam optimizer and cross-entropy loss function were applied. The training lasted 400 epochs.

\subsection{Obtained Results}
The results in terms of confusion matrices on KMU-FED and KDEF datasets are depicted in Figure~\ref{fig:cnx}, showing detailed overview of the recognition performance of our proposed model across different facial expressions. In the case of the KMU-FED dataset, we conducted experiments with varying data splits to investigate the generalization ability of our model. The results showed good performance for most expression categories across different split strategies. When 10-fold cross-validation (Figure~\ref{fig:cnx}a) is used, our model performed very well at identifying surprise, sadness, happiness, and fear. However, the performance decreased at differentiating between the disgust and anger emotions, due to the similarities in their appearance. In 5-fold cross-validation scenario,  excellent performances were obtained by our model, successfully classifying the expressions of fear and surprise with no error rate. Moreover, our model showed very good accuracy in categorizing all expressions while using an 80\%–20\% data split, demonstrating the effectiveness of our proposed model on the KMU-FED dataset. For the KDEF dataset (Figure~\ref{fig:cnx}b), our model performed well, especially in recognizing neutral and happy expressions. However, a slight decrease in performance was observed for expressions of afraid, indicating that it can be challenging to distinguish between expressions of afraid, sad or surprise.

\begin{table*}[htbp]
\caption{Results of the ablation study on KMU-FED dataset. "w" denotes "with"; "w/o" denotes without; and "C" denotes our designed classifier}
\vspace {-2mm}
\label{tab:abstd}
\normalsize
\begin{center}
\begin{tabular}{|c|c|c|c|c|c|c|}
\hline
\textbf{Methods}&{\textbf{Params(M)}}&{\textbf{Processing Time(ms)}}&\textbf{Accuracy}&\textbf{Precision}&\textbf{Recall}&\textbf{F1-score}\\
\hline
ShuffleNet V2 w/o C &2.3M&\textbf{1.03 ms}&93.455\%&0.94&0.93&0.93 \\
\hline
ShuffleNet V2 w C&\textbf{2.0M}&1.11 ms&90.455\%&0.90&0.91&0.90 \\
\hline
EfficientViT-M2 w/o C&4.2M&2.80 ms&86.273\%&0.86&0.86&0.86  \\
\hline
EfficientViT-M2 w C &4.7M&2.70 ms&95.727\%&0.96&0.96&0.96 \\
\hline
\textbf{ShuffViT-DFER (Ours)}&5.9M&3.30 ms&\textbf{97.273\%}&\textbf{0.97}&\textbf{0.97}&\textbf{0.97} \\
\hline
\end{tabular}
\label{tab3}
\end{center}
\end{table*}
\begin{table*}[htbp]
\caption{Comparison between the proposed method and state-of-the-art and existing methods on the KMU-FED dataset}
\vspace {-2mm}
\label{tab:sota1}
\normalsize
\begin{center}
\begin{tabular}{|c|c|c|c|c|}
\hline
\textbf{Methods}&{\textbf{Accuracy (10 Fold)}}&\textbf{Accuracy (5 Fold)}&\textbf{Accuracy (80:20)}\\
\hline
Hierarchical WRF~\cite{jeong2018driver}& 94.7\%&NAN&NAN\\
\hline
CCNN~\cite{zhang2019detecting}& 97.3\% &NAN&NAN \\
\hline
LMRF~\cite{jeong2020lightweight}& NAN &95.1\%&NAN\\
\hline
d-RFs~\cite{jeong2020lightweight}& NAN &91.2\%&NAN\\
\hline
FTDRF~\cite{jeong2020lightweight}& NAN &93.6\%&NAN\\
\hline
DF~\cite{jeong2020lightweight}& NAN &90.5\%&NAN\\
\hline
SqueezeNet~\cite{sahoo2022deep}& 83.4\% & 82.7\% &95.83\%\\
\hline
CNN+SVM~\cite{sukhavasi2022hybrid}& NAN & NAN & 98.64\% \\
\hline
\textbf{ShuffViT-DFER (Our method)} & \textbf{97.3\%} &\textbf{95.6\%} &\textbf{100.00\%} \\
\hline
\end{tabular}
\label{tab1}
\end{center}
\end{table*}
\begin{table}[htbp]
\caption{Comparison between the proposed method and state-of-the-art and existing methods on the KDEF dataset}
\vspace {-2mm}
\label{tab:sota2}
\normalsize
\begin{center}
\begin{tabular}{|c|c|}
\hline
\textbf{Methods}&{\textbf{Accuracy}}\\
\hline
MPCNN~\cite{liu2018multi}& 86.90\%\\
\hline
SquezeeNet~\cite{sahoo2022deep}& 86.86\%  \\
\hline
RBFNN~\cite{mahesh2021shape}& 88.80\% \\
\hline
Efficient-swishNet DCNN~\cite{dar2022efficient}& 85.50\% \\ 
\hline
EfficientNet\_FER~\cite{kalsum2023novel}& 88.17\% \\ 
\hline
\textbf{ShuffViT-DFER (Our method)} & \textbf{92.44\%} \\
\hline
\end{tabular}
\label{tab2}
\end{center}
\end{table}
\vspace {-1mm}
\subsection{Ablation Analysis}
To better gain insight into the performance of our model, we carried out an ablation analysis by comparing the performance of each individual module in our architecture before and after fusion. We considered a variety of factors to ensure a comprehensive comparison including the number of parameters, the processing time of single image, the accuracy, the precision, the recall, and the F1-score. The results are shown in Table~\ref{tab3}. As expected, our model has slightly more parameters and uses longer processing time than the individual modules, while maintaining a relatively low computational cost and providing the best performances, yielding in an average accuracy of 97\% and a processing time of 3.3 ms per a single image after face detection and cropping. The original ShuffleNet V2, either with its original classifier or with our proposed classifier, has less parameters, making it suitable for real-time applications but at the cost of lower performances. When EfficientViT-M2 is considered, the performances are increasing but still are less accurate than our proposed architecture. Based on these results, we can conclude that our proposed method does enhance the performance by combining the strengths of ShuffleNet V2 and EfficientViT-M2 alongside with the proposed classifier. 




\vspace {-1mm}
\subsection{Comparison with State-of-the-Art}
\vspace {-1mm}
We also performed a thorough comparison with state-of-the-art and some recently proposed methods. The results of the comparison are summarized in Tables~\ref{tab1} and~\ref{tab2}. Our approach showed an average accuracy of 97.3\% on the KMU-FED dataset which is comparable to the results published in~\cite{zhang2019detecting} and consistently outperforming all other approaches across different data splits. Our proposed method also outperformed all other methods on the KDEF dataset with an accuracy of 92.44\%. These results assess the validity of our proposed architecture when it comes to the recognition of driver's facial expressions.

\vspace {-4mm}
\section{Conclusion}\label{sec:cn}
\vspace {-1mm}
This paper introduced ShuffViT-DFER, an efficient and fast method for recognizing driver's facial expressions. The approach adopted a transfer learning-based technique, utilizing two lightweight and efficient pre-trained models, ShuffleNet V2 and EfficientViT-M2. The method combines the strengths of the two approaches using a specialized classification scheme. Extensive experiments are conducted on two distinct and challenging datasets simulating real-world scenarios. The obtained results demonstrated the effectiveness of the proposed approach in improving the accuracy while maintaining low processing time as needed in real-time driving environments.

In future work, we plan to explore the integration of multi-modal information, such as combining facial expressions with audio cues. It is also of interest to consider driver's facial expression recognition from multiple cameras to further enhance the performance and cope with the multi-view face angles.

\section*{Acknowledgment} We wish to convey our deep appreciation to the EUNICE Alliance for the financial support.


\bibliographystyle{IEEEtran}
\bibliography{references}
\end{document}